\documentclass[runningheads]{llncs}
\usepackage[T1]{fontenc}

\usepackage{amsmath,amssymb}
\usepackage{graphicx,verbatim}
\usepackage{booktabs}
\usepackage{multirow}
\usepackage{xcolor}
\usepackage{tabularx}
\usepackage{comment}
\usepackage{booktabs,multirow, tabularx,array}
\usepackage[
  colorlinks=true,
  citecolor=blue,
  linkcolor=blue,
    urlcolor=blue
]{hyperref}

\begin{document}
\newcommand{\fb}[1]{{\textcolor{orange}{[Fengbei: #1]}}}
\title{ShapKO: Shapley-Adaptive Modality Knockout for Robust Multimodal Learning}
\titlerunning{ShapKO for Robust Multimodal Learning}
%
\author{Nusrat Binta Nizam\inst{1, 2, 3} \and
Fengbei Liu\inst{1, 3} \and
Sunwoo Kwak\inst{1, 2, 3} \and
Minh Nguyen\inst{1, 2, 3} \and
Ruining Deng\inst{3} \and
Mert R. Sabuncu\inst{1, 2, 3}}
%
\authorrunning{NB. Nizam et al.}
\institute{Cornell University, New York, NY, USA \and
Cornell Tech, New York, NY, USA \and
Weill Cornell Medicine, New York, NY, USA\\
\email{\{nn284,fl453,sk3355,bn244,msabuncu\}@cornell.edu, rud4004@med.cornell.edu}}
  
\maketitle              
\begin{abstract}
Multimodal medical models often degrade when inputs are missing, a common scenario in real-world clinical workflows. Separately, even when all modalities are present, modality dominance is observed during training, where optimization over-relies on a highly predictive modality and undertrains complementary sources, resulting in poor robustness under partial availability. While training-time modality knockout improves missing-modality robustness, existing approaches use static masking rates that cannot adapt to evolving modality utility during training. We introduce ShapKO (Shapley-Adaptive Modality Knockout), a dynamic training strategy that learns modality-specific knockout probabilities based on validation utility. ShapKO periodically evaluates performance across modality subsets, estimates modality importance via Shapley values, and updates masking probabilities to suppress dominant modalities more frequently. This adaptive process promotes complementary representations, while requiring no architectural modifications. We evaluate ShapKO on three datasets covering multitask clinical classification, survival prediction, and cancer detection. ShapKO consistently improves performance under modality absence and yields interpretable trajectories of learned masking behavior. Code is available at: \href{https://github.com/sumona00/ShapKO}{\texttt{https://github.com/sumona00/ShapKO}}.



\end{abstract}

\section{Introduction}
Multimodal learning aims to improve clinical decision support by combining complementary evidence from heterogeneous sources such as imaging, physiological signals, laboratory measurements, and structured clinical variables. In practice, modalities are often missing or corrupted due to cost and triage, conditional ordering, acquisition failures, and site-specific protocols~\cite{wu2024deep}. Consequently, models trained under the assumption that all modalities are present at inference can degrade substantially under realistic missingness~\cite{neverova2015moddrop}.


A key contributor is \emph{modality dominance}: when one modality is consistently most predictive during training, optimization over-relies on it while undertraining weaker but complementary modalities, leading to substantial performance degradation when the dominant modality is unavailable~\cite{javaloy2022mitigating,yang2024facilitating}. Addressing this imbalance requires maintaining performance across diverse modality-availability patterns without training separate subset-specific models~\cite{li2023makes}. Existing robustness strategies, including modality masking, knockout, and gradient balancing, attempt to mitigate this issue by simulating missingness or reweighting modality contributions during training~\cite{nguyen2025knockout,wu2024deep,novosad2024task,mohta2021investigating}. However, these approaches typically rely on static heuristics or local training signals and do not explicitly optimize performance across the combinatorial space of modality subsets. As a result, they cannot adapt to evolving modality utility dynamics and may over-regularize weaker modalities or insufficiently suppress dominant ones. 

Missing modalities are common in clinical data and can substantially degrade multimodal performance, motivating methods that operate under arbitrary availability patterns~\cite{wu2024deep}. In medical imaging, prior work includes any-subset fusion in a shared latent space for inference from partial modality sets~\cite{havaei2016hemis} and generative/shared-latent formulations that support prediction and modality completion~\cite{dorent2019hetero,wu2018multimodal}. Common baselines impute missing inputs or train subset-specific models, but imputation can introduce distribution shift and subset models scale poorly with modality count~\cite{enders2018missing}. A widely used, architecture-agnostic alternative is training-time modality masking/knockout~\cite{nguyen2025knockout,neverova2015moddrop,liu2017learning}, yet static masking rates can exacerbate modality dominance and related collapse phenomena linked to optimization effects such as gradient conflict~\cite{javaloy2022mitigating}. Beyond masking, modality-balancing approaches (e.g., OGM-GE gradient modulation) and configuration-aware models (e.g., task-conditional experts) improve robustness across modality subsets~\cite{peng2022balanced,novosad2024task,guo2024classifier,li2023boosting}.

We propose ShapKO, which learns per-modality knockout probabilities from a validation \emph{utility} evaluated on modality subsets. While standard signals such as validation loss or accuracy on the \emph{all-modality} input provide a single-configuration view that can be dominated by the strongest modality,
our goal is robustness across the many availability patterns encountered at deployment~\cite{zhang2024multimodal,kontras2024improving}. We therefore estimate each modality's average marginal contribution across subsets using Shapley values, a principled coalition-based attribution that accounts for redundancy and complementarity between modalities~\cite{winter2002shapley}. ShapKO periodically evaluates utilities over subsets, estimates modality importance, and updates knockout probabilities with a stable \emph{knockout-strong-more} rule that masks more influential modalities more often, reducing modality dominance and improving robustness under missingness. 
Our contributions are as follows:
\begin{itemize}
    \item We propose ShapKO, a utility-driven, architecture-agnostic modality \emph{knockout} method that 
    adapts per-modality knockout probabilities during training based on Shapley-estimated marginal contributions. To our knowledge, ShapKO is the first method that integrates Shapley-values into training for optimization.
    \item  We evaluate ShapKO in three settings - multitask clinical classification, multimodal prostate cancer detection, and survival prediction - showing improved robustness compared to standard training, fixed-rate knockout (Fixed KO)~\cite{nguyen2025knockout}, and gradient based modulation (OGM-GE)~\cite{peng2022balanced}, providing interpretable knockout-rate trajectories over training.
\end{itemize}


\begin{figure*}[t!]
\includegraphics[width=1.0\textwidth]{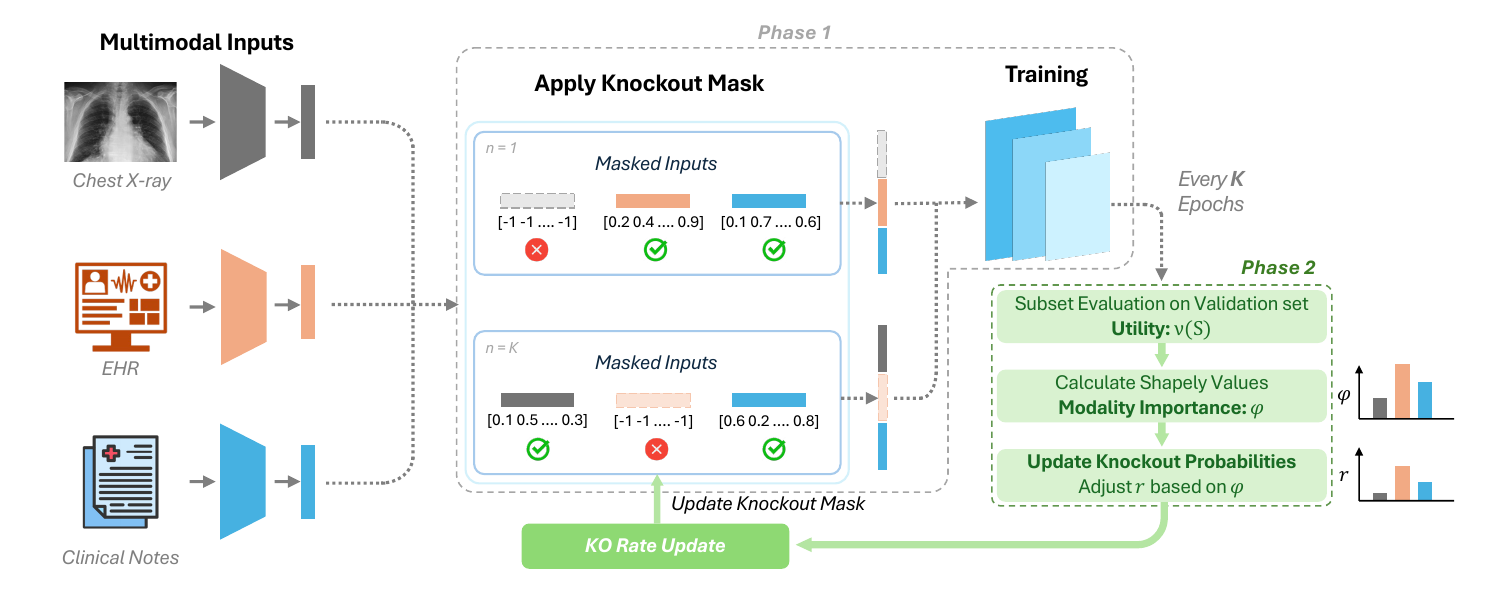}
\caption{\textbf{Overview of ShapKO (architecture-agnostic).}
ShapKO trains with sampled modality masks by replacing knocked-out modalities (or embeddings) with fixed placeholders before fusion. Every $K$ epochs, it evaluates validation utility over modality subsets, estimates Shapley-based modality importance, and updates per-modality knockout probabilities with a stable \emph{drop-strong-more} rule. The procedure is architecture-agnostic and applies across tasks.}
\label{fig1}
\end{figure*}

\section{Methodology}
\label{sec:problem_setup}
\paragraph{\textbf{Problem setup.}} Let $\mathcal{M}=\{1,\dots,d\}$ be the full modality index set and
$\mathcal{O}\subseteq\mathcal{M}$ the modalities available in the dataset. Given a training sample $(x^i, y^i)$, each observed modality $m\in\mathcal{O}^i\subseteq\mathcal{O}$ is encoded by a modality-specific
encoder $g_m$ into an embedding $e^i_m = g_m(x^i_m)\in\mathbb{R}$,
giving $e^i = \{e^i_m\}_{m\in\mathcal{O}^ i}$. ShapKO aims to learn a predictor $f_\theta$ that is robust under arbitrary modality availability during inference. To achieve this, ShapKO introduces additional synthetic knockout during training: a random subset $\mathcal{A}^i\subseteq\mathcal{O}^i$ of present modalities
is further masked independently of $(x^i, y^i)$. 

Modalities in $\mathcal{M}\setminus\mathcal{O}^i$ are \emph{structurally missing} due to
heterogeneous clinical acquisition. and modalities in $\mathcal{O}^i\setminus\mathcal{A}^i$ are \emph{synthetically absent} due to training-time knockout. Both missingness are handled by replacing the corresponding embeddings with different fixed placeholders $\tilde{e}_m$, where the choice of $\tilde{e}_m$ are given in Sec.~\ref{sec:experiments}.

%

\subsection{Utility Estimation and Knockout Rate Update}
\label{sec:utility}
\paragraph{\textbf{Utility function.}} We treat modalities in $\mathcal{M}$ as players in a cooperative
game~\cite{winter2002shapley}. For any subset $S\subseteq\mathcal{O}$, we
define a scalar utility $v(S)$ by evaluating the current model on the
validation set using only modalities in $S$, with all $m\notin S$ replaced by
$\tilde{e}_m$.
Let $J(S)$ denote the task-specific evaluation metric. We define $v(S) = J(S)$ for higher-is-better metrics (e.g., AUC, Accuracy, C-index) and $v(S)=-J(S)$ for lower-is-better metrics (e.g., validation loss), so that larger $v(S)$ always indicates better performance regardless of metric direction. The Shapley value of modality $m$ is:
\begin{equation}
\phi_m \;=\!
\sum_{S\subseteq\mathcal{O}\setminus\{m\}}
\frac{|S|!\,(|\mathcal{O}|-|S|-1)!}{|\mathcal{O}|!}
\Bigl(v(S\cup\{m\}) - v(S)\Bigr),
\label{eq:shapley}
\end{equation}
where $\phi_m$ is the weighted average marginal gain from adding $m$ across all
coalitions~\cite{winter2002shapley}, capturing both its individual contribution
and interactions with other modalities. A larger $\phi_m$ indicates that
modality $m$ is more dominant. We instantiate $J(\cdot)$ per dataset in
Sec.~\ref{sec:experiments}. 
\paragraph{\textbf{Shapley-Adaptive Knockout Rates.}}
We convert Shapley values to weights on the probability simplex by
shifting to non-negative values and normalizing:
\begin{equation}
w_m = \frac{\phi_m - \phi_{\min}}
           {\sum_{m'\in\mathcal{O}}\phi_{m'} - \phi_{\min}},
\quad \phi_{\min} = \min_{m\in\mathcal{O}}\phi_m,
\label{eq:weights}
\end{equation}
so that $w_m \geq 0$ and $\sum_{m\in\mathcal{O}} w_m = 1$.
Let $\bar{w} = 1/|\mathcal{O}|$ be the uniform weight.
Per-modality knockout probabilities are then updated as:
\begin{equation}
    \begin{split}
        r_m &= \mathrm{clip}\!\left(r\!\left(\frac{w_m}{\bar{w}}\right)^{\!\delta},\,
            r_{\min},\, r_{\max}\right), \\
        (1 - r)^d &= 0.5 \;\Rightarrow\; r = 1 - 0.5^{1/d},
    \end{split}
\label{eq:adaptive_rm}
\end{equation}
where $\delta\geq 0$ controls sensitivity (lower for higher structural missingness). $r$ is the base knockout rate initialized to random based on number of modalities. $[r_{\min}, r_{\max}]$ bounds
the rates for stability. Modalities with $w_m > \bar{w}$ (above-average
importance) receive a higher knockout rate than the base $r$, directly
implementing the \emph{drop-strong-more} policy.
\subsection{Training Description}
ShapKO alternates between two phases: optimizing $f_\theta$ under the current
knockout distribution (\textbf{Phase 1}), and updating the knockout rates
based on Shapley-estimated modality importance (\textbf{Phase 2}).

In \textbf{Phase 1}, for each modality $m\in\mathcal{O}_i$ we sample
$z_m \sim \mathrm{Bernoulli}(1-r_m)$ (with $\sum_m z_m \geq 1$ enforced) and
form the knocked-out embedding $\hat{e}_m$,
\begin{equation}
\hat{e}_m =
\begin{cases}
z_m\, e^i_m + (1-z_m)\,\tilde{e}_m, & m \in \mathcal{O}_i, \\
\tilde{e}_m, & m \in \mathcal{M}\setminus\mathcal{O}_i,
\end{cases}
\label{eq:ko_embedding}
\end{equation}
Loss function $\mathcal{L}$ is computed as:
\begin{equation}
    \begin{split}
        \mathcal{L}(\hat{e}^i_m, y^i) = \frac{1}{|\mathcal{O}|}\sum_{m=1}^{|\mathcal{O}_i|} \mathcal{L}\!\left(f_{\theta}(\hat{e}^i_m),\; y^i\right),
    \end{split}
\end{equation}

In \textbf{Phase 2}, triggered every $K$ epochs after
warm-up, the current $f_\theta$ evaluates $v(S)$ over all subsets in validation split,
estimates $\phi_m$ (Eqs.~\eqref{eq:shapley}), and
updates $r_m$ (Eqs.~\eqref{eq:weights}--\eqref{eq:adaptive_rm}), with
$\theta$ fixed.

\section{Experimental Design}
\begin{table}[t]
\centering
\small
\setlength{\tabcolsep}{4pt}
\renewcommand{\arraystretch}{0.9}
\caption{Hyperparameters and placeholder values used across tasks.}
\label{tab:hyperparams}
\resizebox{\columnwidth}{!}{%
\begin{tabular}{@{}lccccccccccc@{}}
\toprule
& Epochs
& \shortstack{Learning\\rate}
& \shortstack{Weight\\decay}
& Optimizer
& $K$
& $d$
& $r_{\min}$
& $r_{\max}$
& $\delta$
& \shortstack{Struct.\\placeholder}
& \shortstack{Synth.\\placeholder} \\
\midrule
Prostate MRI        & 100 & 3e-4 & 1e-4 & AdamW & 15 & 3 & 0.02 & 0.30 & 0.5 & 0              & 0              \\
Survival Prediction & 300 & 2e-4 & 5e-5 & AdamW  & 20 & 4 & 0.02 & 0.60 & 0.5 & 0              & $-1$           \\
Multitask Classif.  & 100 & 1e-4 & 1e-5 & AdamW  & 10 & 3 & 0.02 & 0.35 & 1.0 & \texttt{[MISS]}& \texttt{[KO]}  \\
\bottomrule
\end{tabular}%
}
\end{table}

\label{sec:experiments}

\subsection{Dataset and Implementation details}
\paragraph{\textbf{Dataset organization.}}
We evaluate on three multimodal benchmarks covering complementary clinical tasks. \textbf{FlexCare} integrates structured EHR variables, chest imaging, and clinical text (MIMIC-IV/CXR/NOTE) for multi-task clinical classification~\cite{xu2024flexcare}. For time-to-event modeling, we use the \textbf{MMD} benchmark combining radiology, pathology, genomics, and demographics for survival prediction~\cite{cui2022survival}.
For imaging robustness, we use \textbf{prostate MRI datasets} with T2w, DWI, ADC modalities for binary clinically significant prostate cancer detection~\cite{saha2024artificial}.


\paragraph{\textbf{Implmentation details.}} For Prostate MRI datasets, we use U-Net style 3D CNNs as modality encoder $g_m$. Embeddings are concatenated and fed to a shallow fusion MLP for binary prediction, trained with binary cross-entropy loss~\cite{mannor2005cross}. For $J(\cdot)$ we use AUC as metric. We use a fixed zero placeholder $\tilde{e}_m$ for both structural and synethic missingness. For MMD, we follow previous work~\cite{cui2022survival} and use the same modality-specific encoders for pathology, radiology, genomics, and demographics. We use C-index for $J(\cdot)$ due to standard metric for survival prediction. We adopt a fixed zero placeholder for structural missingness and -1 placeholder for synthetic missingness. We optimize the model using the Cox partial-likelihood loss as formulated in~\cite{cui2022survival,sasieni1993maximum}. For FlexCare, we follow their setting and adopt the tri-modal Transformer architecture over EHR time-series, chest X-ray (CXR) images, and clinical notes~\cite{xu2024flexcare}. We select averaged multi-task AUC as $J(\cdot)$.  We use a learned \texttt{[MISS]} token embedding for structural missingness and a separate \texttt{[KO]} token embedding for synthetic missingness, with attention masking to prevent information leakage under partial availability~\cite{shi2023mftrans}. Note we drop the Mixture of Experts (MoE) component in the original FlexCare implementation for fair comparison with other baselines. We train this model with a task-specific supervised loss (binary cross-entropy for most tasks and cross-entropy for multi-class tasks such as length-of-stay)~\cite{mannor2005cross}.

\paragraph{\textbf{Hyperparameters.}} Table~\ref{tab:hyperparams} provides the hyperparameters and placeholder values for different tasks. All hyperparameters were set empirically and refined by cross-validation on the training folds, selecting values that yielded stable learning dynamics and the best validation performance.

\section{Results and Discussions}
\begin{figure*}[t!]
    \centering
    \includegraphics[width=0.9\linewidth]{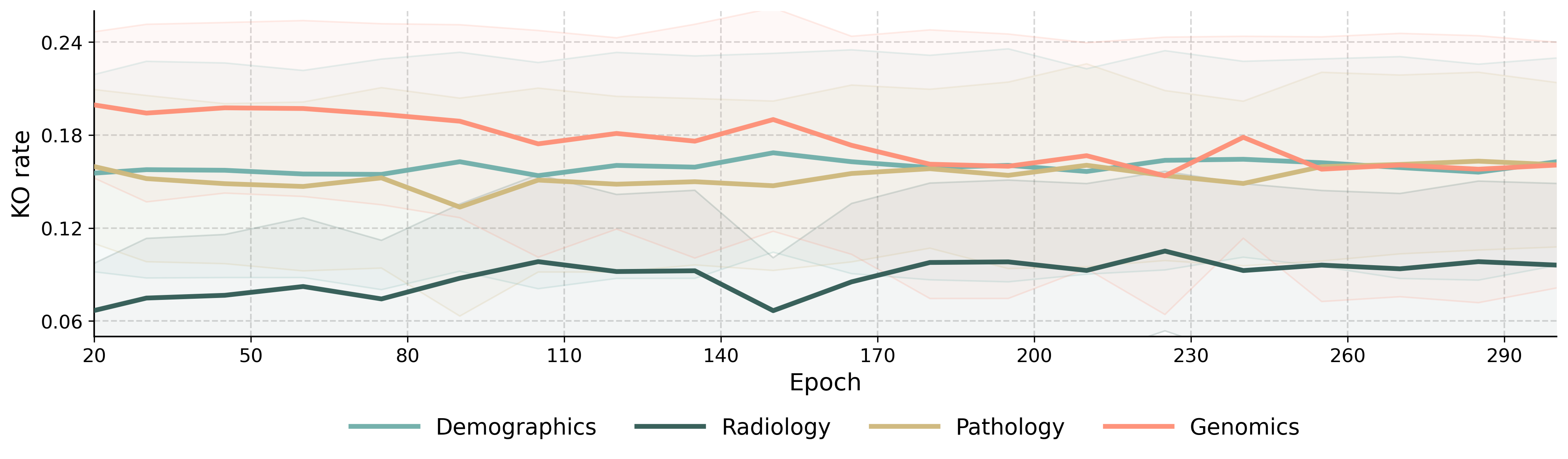}
    \caption{ShapKO rates over training (warm up epoch, K =20). Mean (solid line) and standard deviation (shaded band) of the learned adaptive knockout rate for each modality across 15 cross-validation folds, plotted at the KO update epochs. Lower KO indicates the modality is retained more often, while higher KO indicates more frequent knockout.}
    \label{fig:adaptive_ko_rates}
\end{figure*}

\begin{figure}[t!]
  \centering
  \includegraphics[width=\textwidth]{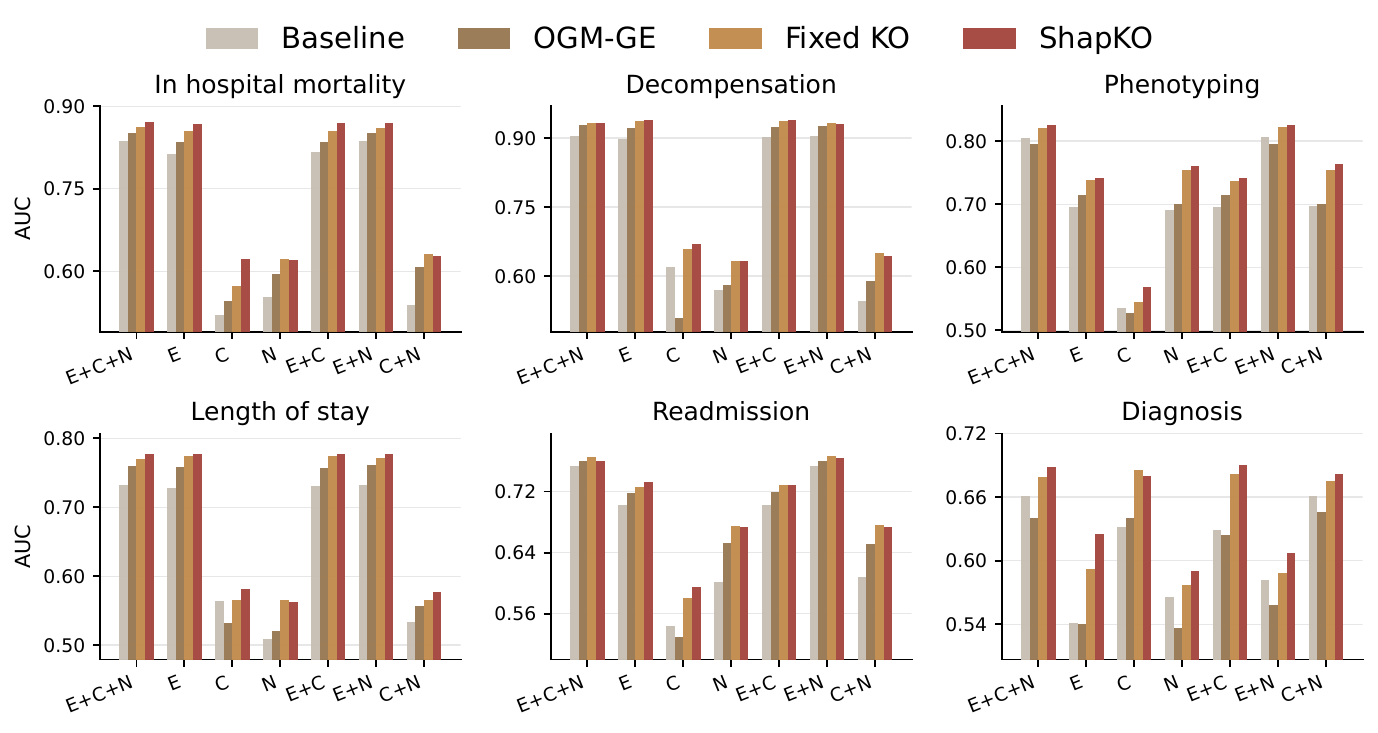}
  \caption{Impact of modality knockout on MIMIC-IV multitask classification. Grouped bar plots report AUC for six clinical prediction tasks (In-hospital mortality, Decompensation, Phenotyping, Length of stay, Readmission, and Diagnosis) under different input modality subsets. Modality shorthand: E = EHR, C = chest X-ray, N = clinical notes (e.g., E+C+N uses all modalities). We compare Baseline training, OGM-GE, Fixed KO, and ShapKO across all subsets, showing consistent gains for ShapKO, particularly under missing-modality settings.}
  \label{fig:mimic_auc_grid}
\end{figure}

\begin{table}[t!]
\centering
\small
\setlength{\tabcolsep}{6pt}
\renewcommand{\arraystretch}{1.1}
\caption{AUC (\%, mean$\pm$std over 5 folds) for binary classification under different modality combinations. Best method per row is bolded.}
\label{tab:picai_modality_combos_auc}
\begin{tabular}{lcccc}
\toprule
\textbf{Modality subset} & \textbf{Baseline} &  \textbf{OGM-GE} & \textbf{Fixed KO} &\textbf{ShapKO} \\
\midrule
T2w                & 0.656$\pm$0.127 & 0.636 $\pm$ 0.031 & \textbf{0.685$\pm$0.048} & 0.663$\pm$0.067 \\
ADC                & 0.732$\pm$0.084 &0.726 $\pm$ 0.065 & 0.738$\pm$0.095 & \textbf{0.739$\pm$0.143} \\
DWI                & 0.773$\pm$0.157 &0.787 $\pm$ 0.041 & 0.818$\pm$0.055 & \textbf{0.842$\pm$0.055} \\
\midrule
T2w+ADC            & 0.747$\pm$0.082 & 0.724 $\pm$ 0.067 & 0.765$\pm$0.070 & \textbf{0.775$\pm$0.113} \\
T2w+DWI            & 0.777$\pm$0.150 & 0.787 $\pm$ 0.040 & 0.824$\pm$0.045 & \textbf{0.855$\pm$0.036} \\
ADC+DWI            & 0.831$\pm$0.025 & 0.819 $\pm$ 0.018 & 0.839$\pm$0.040 & \textbf{0.878$\pm$0.038} \\
\midrule
T2w+ADC+DWI        & 0.831$\pm$0.026 & 0.818 $\pm$ 0.017 & 0.840$\pm$0.038 & \textbf{0.887$\pm$0.029} \\
\bottomrule
\end{tabular}
\end{table}
\begin{figure*}[h!]
    \centering
    \includegraphics[width=\linewidth]{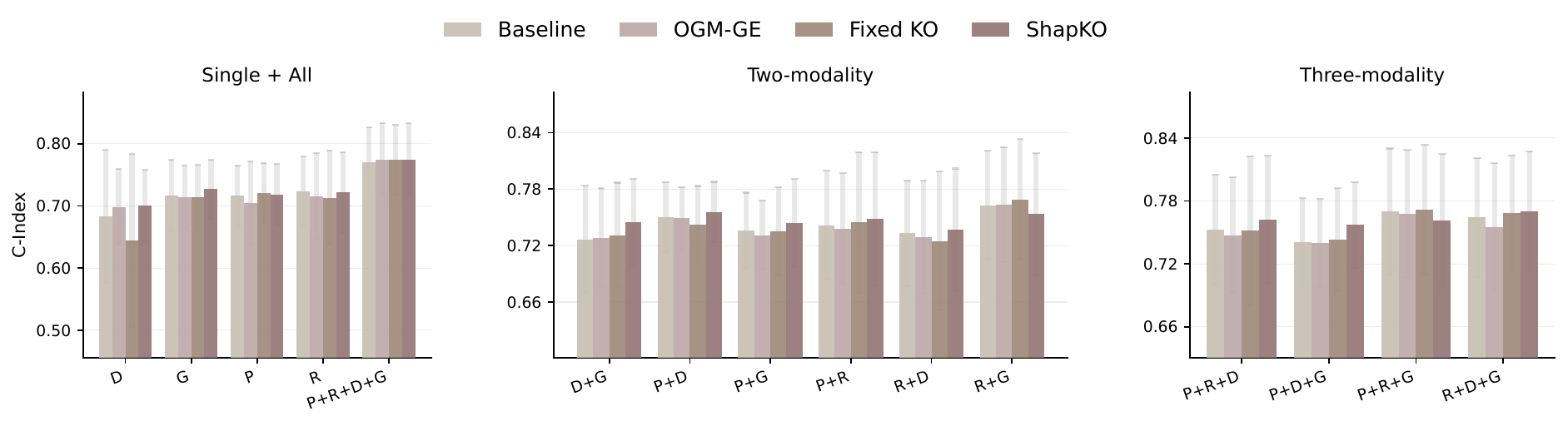}
    \caption{C-index performance across modality subsets under different strategies (across 15 cross-validation folds). Grouped bar plots compare Baseline, OGM-GE, Fixed KO, and ShapKO for single-modality + full-modality, two-modality, and three-modality input subsets. Vertical gray ranges indicate the variability (± standard deviation) around each estimate. D = Demographics, G = Genomics, P = Pathology, R = Radiology. Combinations denote modality fusion, e.g., P+R = Pathology + Radiology, and P+R+D+G = all four modalities.}
    \label{fig:mmd_bar}
\end{figure*}
Across all benchmarks, modality knockout improves robustness to missing inputs by reducing train--test mismatch and mitigating modality dominance. In particular, ShapKO consistently yields the strongest performance under partial modality availability, indicating that utility-driven knockout can better calibrate reliance on modalities as training progresses. 
\paragraph{\textbf{Prostate Cancer Detection.}}
Table~\ref{tab:picai_modality_combos_auc} summarizes binary classification AUC under all modality combinations of T2w, ADC, and DWI. Fixed KO increases AUC over the baseline and OGM-GE for several subsets, while ShapKO achieves the best performance for most combinations. In particular, ShapKO achieves the highest AUC for all the combinations except T2w. These results indicate that ShapKO strengthens performance when all modalities are available and yields consistent gains when deployed inputs are incomplete, a frequent scenario in prostate MRI due to missing or corrupted sequences. 
\paragraph{\textbf{Survival Prediction.}}
Figure~\ref{fig:mmd_bar} compares C-index values across all modality subsets (D,G,P,R). Fixed KO improves over the baseline in many missing-modality settings, and OGM-GE provides less
consistent gains.  ShapKO achieves performance that is consistently comparable to or better than Fixed KO across most modality subsets.
While the absolute improvements in C-index are modest, ShapKO avoids degradation in any configuration and provides more stable performance when prediction relies on less informative modality combinations. 
Importantly, ShapKO maintains baseline level performance in the full-modality setting while improving robustness under partial-modality evaluation, indicating that this training approach does not compromise overall survival modeling capacity.
Figure~\ref{fig:adaptive_ko_rates} shows non-uniform ShapKO knockout-rate trajectories (mean$\pm$std), indicating adaptation to modality utility. The knockout-rate trajectories provide a simple diagnostic of modality utility drift over training, revealing non-uniform knockout behavior across modalities.
\paragraph{\textbf{Multitask Classification.}}
Figure~\ref{fig:mimic_auc_grid} reports AUC for six clinical prediction tasks under modality subsets of
EHR (E), chest X-ray (C), and clinical notes (N). Both Fixed KO and ShapKO improve over the baseline and
OGM-GE in missing-modality settings, showing that training with modality knockout yields models that
generalize better to partial inputs. ShapKO is consistently strongest across subsets, with especially
clear gains in single-modality inference (E-only, C-only, or N-only), where robustness depends on how well each unimodal encoder is trained despite multimodal optimization. These results are consistent with
ShapKO mitigating modality dominance by adaptively increasing knockout on influential modalities over
training, encouraging better use of complementary signals when other modalities are absent.

\section{Limitations and Future Work}
As future work, the validation-time overhead of ShapKO could be reduced by replacing exhaustive subset evaluation (including task-specific metrics) with approximate Shapley estimation, e.g., subset sampling,
permutation-based Monte Carlo, or surrogate explainers such as KernelSHAP~\cite{witter2025regression,mitchell2022sampling}, while preserving accurate modality-importance tracking. ShapKO also depends on design choices such as placeholder values, the subset used for utility estimation, and hyperparameters, which may affect stability and require tuning across datasets. Finally, ShapKO optimizes average subset utility and may not explicitly prioritize worst-case or clinically prevalent missingness patterns; future work could incorporate distribution or risk-aware update rules and learn more expressive missing/knockout embeddings~\cite{uddin2025hybrid}.
\section{Conclusion}
We presented ShapKO, a Shapley-guided modality knockout strategy that dynamically adjusts knockout rates to reduce modality dominance and improve robustness to missing inputs. Across multitask classification, survival prediction, and prostate cancer detection, ShapKO consistently improved performance under partial-modality settings and often matched or exceeded full-modality baselines. These results highlight the value of utility-driven, training-time modality knockout for reliable multimodal learning in real-world clinical deployments. Moreover, ShapKO yields interpretable knockout-rate trajectories that provide a direct diagnostic of evolving modality utility and reliance during training.

 \bibliographystyle{splncs04}
 \bibliography{mybibliography}
\end{document}